# A Non-Local Conventional Approach for Noise Removal in 3D MRI

Sona Morajab[1,*] mehregan Mahdavi[2]

[1]Faculty of engineering, Electronic department, Ayandegan Institute of Higher Education, Tonekabon, Iran, *morajab@aihe.ac.ir*

[2] School of Computer Science and Engineering, The University of New South Wales, Australia. *mehrm@cse.unsw.edu.au*

## ABSTRACT

In this paper, a filtering approach for the 3D magnetic resonance imaging (MRI) assuming a Rician model for noise is addressed. Our denoising method is based on the Conventional Approach (CA) proposed to deal with the noise issue in the squared domain of the acquired magnitude MRI, where the noise distribution follows a Chi-square model rather than the Rician one. In the CA filtering method, the local samples around each voxel is used to estimate the unknown signal value. Intrinsically, such a method fails to achieve the best results where the underlying signal values have different statistical properties. On the contrary, our proposal takes advantage of the data redundancy and self-similarity properties of real MR images to improve the noise removal performance. In other words, in our approach, the statistical momentums of the given 3D MR volume are first calculated to explore the similar patches inside a defined search volume. Then, these patches are put together to obtain the noise-free value for each voxel under processing. The experimental results on the synthetic as well as the clinical MR data show our proposed method outperforms the other compared denoising filters.

## 1. INTRODUCTION

Raw Magnetic Resonance (MR) data are complex valued and corrupted by additive white noise[1]. The main sources of noise include thermal and inductive losses [2]. Such a noise can degrade the visual quality and diagnosis ability notably. Furthermore, the successful performance of many other computerized algorithms such as segmentation and registration depends on the raw signal being as less noisy as possible [3].

Due to the physiological and anatomical concerns, the final MR image is formed by calculating the magnitude of the raw MR data[2]. It is shown that this non-linear process will change the distribution of noise to the Ricain random probability density function[4]. Hence, two specific properties of the MR data should be considered while developing MR denoising filters: 1. the anatomical structures of images including small details must be kept when suppressing the noise fluctuation 2. The Rician nature of noise has to be taken into account. Note that, the estimation of the noise-less signal value becomes more complex in the presence of the Ricain noise, specifically in low SNR values where the signal-dependent bias produced by noise is a difficult issue to deal with [4, 5].

So far, a verity of denoising filters has been proposed to remove noise in MRI. McGibney and Smith [5] presented the first attempts to eliminate the noise impact from MR images. Later on, many other traditional image denoising approaches such as total variation [6] and anisotropic diffusion [7, 8] have been developed to properly work with the MR data. Several denoising methods have been presented based on the well-known non local means (NLM) filter [9] such as optimized blockwise NLM (OBNLM) [10-12], unbiased NLM (UNLM) [13, 14], and pre-filtered rotationally invariant NLM (PRINLM) [15]. In contrast to the original NLM, the aforementioned methods consider the Ricain nature of noise in MRI, leading to more efficient filtering performance.

Another class of the MRI denoising filters includes different kinds of statistical estimation methods. Various realizations of the Maximum Likelihood (ML) [16-18] estimation have been suggested for MR restoration purposes. A non-parametric estimation approach depending on zeroth and second orders 3D kernel regression has been presented by Lopez-Rubio *et al* [19]. Aja-Fernandz [20, 21] employed the linear minimum mean square error (LMMSE) to deal with the Rician distributed random variables. Considering the redundancy of data in the real MR data, a non local LMMSE approach has been proposed in [22, 23], where the Bayesian error between the estimated and the ground-truth signal is tried to be minimized, resulting in outstanding denoising capacity.

Furthermore, transform domain filtering approaches have extensively been used for MRI denoising goals. Many noise removal methods are given based on wavelet and wavelet packet theory [24-26]. In addition, some DCT based filters, e.g., Oracle based DCT, have been presented that reach state-of-the-art results [13].

In this paper, we develop a new MRI denoising method based on the Conventional Approach (CA) [5]. This filter has a closed-form solution for the Rician distributed data. Therefore, it takes advantage of the low computational cost. However, it applies a local processing approach to calculate the noise-less signal value for each voxel under consideration. As shown in many literature [12-15, 27, 28], such a local implementation does not benefit from the whole capacity of a redundant dataset, e.g. MRI. In addition, it leads to sub-optimal results when the underlying signal comes from different distribution (i.e., the difference between the underlying grey levels in a local neighbor is large). In this paper, using the self-similarity property of the patches in a redundant field like MR data, a non local extension of the CA filter is presented, which addresses the aforementioned drawback of the original CA filter.

The rest of this paper is organized as follows: The proposed method is elaborated in Section 2. Section 3 gives the quantitative and qualitative results for several MRI denoising filters using both synthetic and real MR dataset. Conclusion and some remarks are given in Section 4.

## 2. METHODOLOGY

### 2. 1. Proposed Method

Since the proposed method is based on the Conventional Approach (CA) [5], it is worth mentioning the main properties of this method here. Considering the relation between noise and signal of the second order moment in a Ricain distribution, the underlying signal can be estimated as follows [20]:

$$\hat{A} = \sqrt{\max(\langle M^2 \rangle - 2\sigma_n^2, 0)} \qquad (1\text{-}a)$$

$$\langle M^2 \rangle = \frac{1}{N} \sum_{i \in N} M_i^2 \qquad (1\text{-}b)$$

where, $\hat{A}$ and $M$ are respectively the estimated noise-less signal value and noisy Ricain distributed MR data. $\sigma_n$ is the standard deviation of the underlying Gaussian noise. And, $\langle \cdot \rangle$ is the sample estimator given by Eq. (1-b).

As mentioned earlier, to calculate the sample estimator using the CA filter, the intensity values within a local 3D neighbor around each voxel under processing is used as the samples. However, it has been proven that this is not an efficient approach to deal with this problem [27, 28], as many similar samples can be found throughout the dataset that are not necessarily located inside a small local neighbor [28]. Moreover, the assumption of having similar statistical property fails when the noise-free signal values are not the same (e.g., over the edges). To resolve this drawback, a Non-Local Conventional Approach (NLCA) is developed here. Our approach aims to find the samples with similar statistical properties through a search 3D volume defined around each voxel of the given MR data. As shown in [16], the first and second statistical momentums are reliable measures to evaluate the similarity of samples. These measures are defined as follows:

$$C_1 \leq \frac{\langle M_n \rangle}{\langle M_c \rangle} \leq \frac{1}{C_1} \quad and \quad C_2 \leq \frac{\langle M_n^2 \rangle}{\langle M_c^2 \rangle} \leq \frac{1}{C_2} \tag{2}$$

where, $C_1$ and $C_2$ are two constants in the range of 0 and 1 that control the strictness of the sample selection measure (In the experiments, it appeared that $C_1 = 0.9$ and $C_2 = 0.5$ can lead to the best denoising performance with our denoising approach). The sub-index *c* and *n* respectively represent the patch (i.e., a 3D neighbor of size $3^3$) under consideration and the other patches to be compared with it. Other variables are defined exactly the same as those given for Eq. (1).

Even though, Eq. (2) demands to search the whole 3D MR volume to find the similar patches, this process is computationally too expensive. Hence, as given in many literatures [11-15], a search sub-volume around each voxel under-processing is considered to find the similar patches. In our experiments, it turned out that a search volume of the size $11^3$ can be a relevant trade-off between the computational cost and denoising performance.

Substantially, Eq. (2) evaluates the similarity of samples in two different positions of the MR data by comparing their mean and variance values defined inside a patch. It means the samples that cannot meet the criteria imposed by Eq. (2) are discarded from the estimation process. In this way, the similar patches are selected not only from the local neighborhood around the desired voxel but also from a non-local neighborhood within the search sub-volume. This leads to a more efficient sample selection approach that just includes the samples with more similar statistical properties. Furthermore, discarding the non-similar samples intrinsically ameliorates the estimation performance.

**2. 2. Estimation of the Noise Variance**

Having an accurate estimation of the noise level is a necessary step in many denoising filters applied on real MR images. Hence, we employ a robust Rician noise estimation proposed by [29] to obtain the noise level before applying any denoising filters. We picked this method as it is well-matched to Rician distribution and 3D MR data. In contrast to many existing noise estimation methods that need to crop a region from background or signal to estimate the noise level, this method helps relax the assumptions performed by background-based methods (i.e., Rayleigh noise approximation) and signal-based methods (i.e., Gaussian noise approximation). This method applies a MAD estimator [30] on the 3D high sub-band (HHH) of the wavelet coefficients that enables the noise estimation in the presence of the Gaussian distributed random variable. Then, it uses a correction scheme to do the Ricain adaption and obtain an unbiased estimation of the noise level for all signal-to-noise (SNR) values [29]:

$$\hat{\sigma} = \frac{median(|y_i|)}{0.6745} \qquad (3\text{-}a)$$

$$\hat{\sigma}_n = \sqrt{\hat{\sigma}^2 / \xi(\theta)} \qquad (3\text{-}b)$$

$$\xi(\theta) = 2 + \theta^2 - \frac{\pi}{8} \exp\left(-\frac{\theta^2}{2}\right)\left((2+\theta^2) I_0\left(\frac{\theta^2}{4}\right) + \theta^2 I_1\left(\frac{\theta^2}{4}\right)\right)^2 \qquad (3\text{-}c)$$

where, $y_i$ is the wavelet coefficient of the HHH sub-band and $\hat{\sigma}$ is the estimation of noise in the magnitude image. $\theta$, $I_0$, and $I_1$ are respectively the SNR value, first, and second order modified Bessel functions. Note that, $\hat{\sigma}_n$ is the underlying standard deviation of the Gaussian noise needed for denoising filters. A block diagram of the proposed method is shown in Fig. 1.

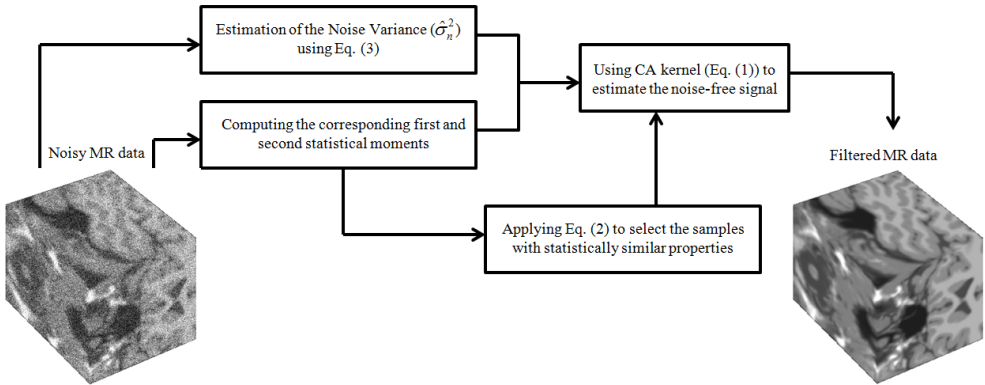

**Fig. 1.** The block diagram of the proposed method.

## 3. EXPERIMENTS and RESULTS

### 3. 1. Validation on the Synthetic Dataset

To compare the performance of the proposed NLCA approach against the other compared MRI denoising methods, we employ the well-known noise-free MRI dataset from Brainweb [31] including a T1w, T2w, and PDw 3D MR volumes of the size $181 \times 217 \times 181$ voxels (voxel resolution=1 mm$^3$, bit precision = 8). Note that, to provide a full evaluation of methods in different noise power, a wide range of the Rician noise (5-20% of the maximum gray level) is generated on the acquired dataset as follows[16]:

$$M = \sqrt{(A+n_1)^2 + n_2^2}. \qquad (4)$$

where $A$ and $M$ are respectively the noise-free and Rician distributed noisy MR data. $n_1$ and $n_2$ are two Gaussian random numbers $n_1, n_2 \sim (0, \sigma_n^2)$.

Root mean squared error (RMSE) and structural similarity index (SSIM) are two widely-used quantitative measures used in the MRI literatures. Hence, we used these measures to compare the denoising ability of the compared methods in this paper. RMSE between the ground-truth signal A and the estimated one Â with the same size N is calculated as follows [28]:

$$RMSE = \sqrt{\frac{1}{N}\sum_{x=1}^{N}\left|A_x - \hat{A}_x\right|^2}. \qquad (5)$$

While the RMSE measures the difference between two dataset in a voxel-by-voxel approach, the SSIM index considers the overall visual similarity. As a result, it is more consistent with the human visual system. The SSIM index is obtained using the following equation [28]:

$$SSIM(A, \hat{A}) = \frac{1}{N}\sum_{x,y=1}^{N}\frac{(2\mu_x\mu_y + c_1)(2\sigma_{xy} + c_2)}{(\mu_x^2 + \mu_y^2 + c_1)(\sigma_x^2 + \sigma_y^2 + c_2)}. \qquad (6)$$

where, $\mu_x$ and $\mu_y$ are respectively the local mean values of dataset $A$ and $\hat{A}$, and $\sigma_x$, $\sigma_y$, and $\sigma_{xy}$ are the corresponding standard deviations and covariance values. $c_1$ and $c_2$ are two constants.

The proposed NLCA method is compared against several MRI denoising including NLMMSE [27], CA [5], UNLM[13], and OBNLM[10]. In our synthetic experiments, the variance of noise is set to its exact value where it is needed, and the filtering methods are applied using the best parameters proposed by their authors. These will lead to the optimal performance for all methods.

Table I compares the filtering methods with the aforementioned synthetic dataset and quantitative measures. As seen, our presented NLCA shows the best performance in almost all cases specifically where the destructive nature of noise in high noise levels decreases the visual quality drastically. This confirms that our proposed approach is suitably fit to the Rician nature of noise. Comparing the performance of the CA and NLCA is an interesting case. As shown, the non-local extension of the CA approach leads to remarkable improvement over the results compared to the CA filter. This in return can prove the presence of data redundancy in the MR data as mentioned in many related literatures [13, 27, 28]. Among the other filters, the results of the NLMMSE and OBNLM show desirable denoising ability. However, their performance dwindles when the noise level exceeds 10%.

Table I. The quantitative comparisons between different MRI denoising filters for the noise level [5%, 20%] and different 3D MR dataset. In each case the best value is highlighted

| | | 5% | | 10% | | 15% | | 20% | |
|---|---|---|---|---|---|---|---|---|---|
| | | RMSE | SSIM | RMSE | SSIM | RMSE | SSIM | RMSE | SSIM |
| T1w | CA | 6.0379 | 0.9270 | 6.6124 | 0.9048 | 7.4904 | 0.8712 | 8.6818 | 0.8262 |
| | NLMMSE | 3.4920 | 0.9671 | 5.2312 | 0.9331 | 6.8815 | 0.9023 | 8.1214 | 0.8552 |
| | OBNLM | **3.3813** | **0.9709** | 5.4475 | 0.9314 | 7.0425 | 0.8888 | 8.4235 | 0.8427 |
| | UNLM | 3.9203 | 0.9677 | 5.1228 | 0.9327 | 6.8807 | 0.8917 | 8.3815 | 0.8530 |
| | NLCA | 3.5501 | 0.9663 | **5.0896** | **0.9384** | **6.4443** | **0.9071** | **8.0398** | **0.8601** |
| T2w | CA | 12.508 | 0.8955 | 13.351 | 0.8393 | 14.820 | 0.7815 | 17.247 | 0.6812 |
| | NLMMSE | 5.6665 | **0.9547** | **8.5657** | 0.8879 | 11.741 | 0.7937 | 14.909 | 0.7308 |
| | OBNLM | 5.5012 | 0.9510 | 8.7535 | 0.8849 | 11.961 | 0.7912 | 15.102 | 0.7345 |
| | UNLM | 5.7119 | 0.9492 | 8.9231 | 0.8775 | 12.004 | 0.7901 | 15.223 | 0.7115 |
| | NLCA | **5.3605** | 0.9485 | 8.6057 | **0.8885** | **11.340** | **0.8092** | **14.871** | **0.7418** |
| PDw | CA | 7.0769 | 0.9294 | 8.4226 | 0.8617 | 10.292 | 0.8023 | 13.481 | 0.7094 |
| | NLMMSE | 4.3159 | 0.9415 | 5.5213 | **0.8881** | 8.0937 | 0.8397 | 10.188 | 0.7327 |
| | OBNLM | 4.5880 | 0.9343 | 6.2509 | 0.8733 | 8.3298 | 0.8055 | 10.210 | 0.7284 |
| | UNLM | 4.7603 | 0.9310 | 6.5479 | 0.8687 | 8.9420 | 0.8215 | 10.312 | 0.7255 |
| | NLCA | **4.2319** | **0.9433** | 5.5063 | 0.8797 | **7.8640** | **0.8431** | 10.332 | **0.7512** |

Fig. 2 provides a visual comparison of different methods for different MR modalities and noise power. As seen, the proposed NLCA removes noise trace suitably while retaining the important vessels and detailed structures. The results of the OBNLM and UNLM are comparable with those of the NLCA. However, these methods typically over-smooth the denoised images and blur some small details while suppressing noise. This is mainly because of the presence of the weighted averaging kernel of the NLM-based filtering methods.

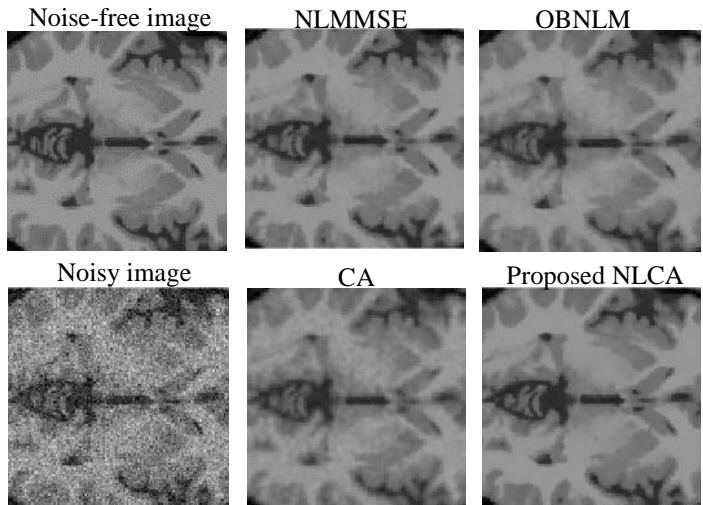

**Fig. 2.** Example denoising results of the filtering methods for 10% of the Rician noise and T1w dataset. A typical axial cut inside the skull is shown in each case.

### 3. 2. Validation on the Clinical Dataset

To evaluate the compatibility of the proposed NLCA filtering approach with clinical MR dataset, we carried out our experiments on different real MRI acquired from a Simens 1.5 T scanner. For the sake of brevity, the result of a T1w 3D MR volume (TR = 500ms, TE = 14 ms, flip angle = 90, volume size= $512 \times 512 \times 36$, voxel resolution = $0.5 \times 0.5 \times 2$ mm$^3$) is just presented here. Since there is no ground-truth available for the real dataset, the filtering performance can only be examined by visual quality of the outputs. The result of the proposed NLCA is shown in Fig. 3. As seen, in addition to a suitable contrast between white and gray matters, the filtered image gives a reliable capability in maintaining the detailed structures and there is no trace of the anatomical structures on the residual image. All the visual assessments have been verified by an expert radiologist and a neurosurgeon.

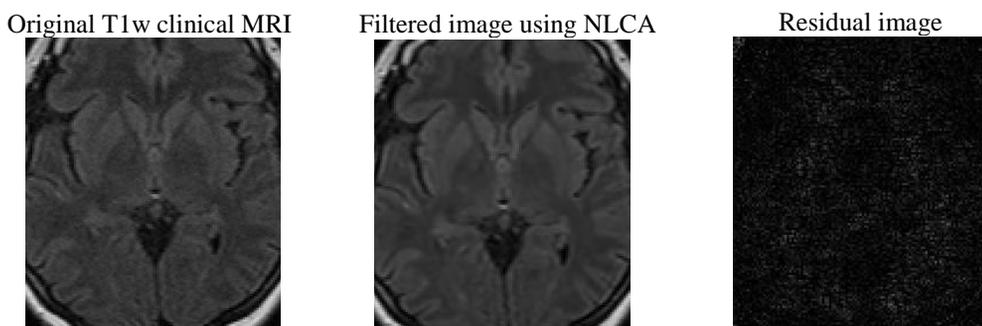

**Fig. 3.** Denoising result of the proposed NLCA filter applied to the real brain T1w MR dataset.

## 4. CONCLUSION and DISCUSSION

In this paper, the data redundancy and self-similarity properties of the 3D MR dataset were used to develop a non-local extension of the Conventional Approach (CA). Our denoising method aimed to deal with Rician noise which is extensively used to model existing noise on the single coil magnitude MRI. With our proposed NLCA method, the selection of sample are not restricted to a local area around each voxel under-processing, but a similarity measure based on the statistical momentums are used to extract the similarly distributed samples inside a non-local neighborhood. The presented method is capable of taking into account both the noise characteristics and image structures. This in return causes remarkable improvement compared to the original CA method.

Different validations were done on both synthetic and real 3D MR data using various image quality measures. The performance of the proposed method was compared against several methods developed to deal with 3D MRI assuming Ricain distributed random variables. Considering the results, the proposed NLCA outperforms the compared filters in almost all cases. In terms of detailed structures, which are of great importance for pathological and diagnosis purposes, the proposed NLCA shows reliable performance; this was verified by an expert radiologist and a neurosurgeon.

Developing new similarity measures can be a suitable extension to this work. Moreover, applying the proposed method on multi-coil MR data, which follows a non central Chi-square noise model, is another interesting future research topic.

## ACKNOWLEDGMENT

We are grateful to Prof. Alizadeh, Poursina Hospital, Rasht, Iran, for providing access to clinical MR dataset and also his valuable comments on the quality of the filtered MR data.